\documentclass[runningheads]{llncs}
\usepackage[T1]{fontenc}
\usepackage{graphicx,verbatim} 

\usepackage{multirow} 
\usepackage[colorlinks=true,allcolors=blue]{hyperref} 
\usepackage[misc]{ifsym}
\usepackage[normalem]{ulem} 
\useunder{\uline}{\ul}{}
\usepackage{amsmath,amssymb,amsfonts,bm} 

\usepackage[table,xcdraw]{xcolor}
\usepackage{floatrow} 
\usepackage{booktabs} 
\usepackage{placeins}
\newfloatcommand{capbtabbox}{table}[][\FBwidth]
\begin{document}
\title{U-TTT: Towards Generalizable PET Image Denoising via Test-Time Training}
\titlerunning{U-TTT}
%
\author{
Zhiwen Yang\inst{1} \and
Jiayin Li\inst{1} \and 
Hao Lu\inst{1} \and 
Hui Zhang\inst{2} \and 
Zihua Wang\inst{3} \and 
Yan Xu\inst{1}$^{(\textrm{\Letter},\thanks{Corresponding author})}$
} 


\authorrunning{Yang et al.}
\titlerunning{U-TTT}
\institute{
School of Biological Science and Medical Engineering, Beihang University, Beijing 100191, China
\\
\email{xuyan04@gmail.com} \and 
Department of Biomedical Engineering, Tsinghua University, Beijing 100084, China \and
School of Aerospace Engineering, Tsinghua University, Beijing 100084, China
}

\maketitle      

\begin{abstract}
Existing deep learning models for Positron Emission Tomography (PET) image denoising often suffer from severe performance degradation under distribution shifts, fundamentally restricting their robust clinical deployment. This lack of generalization stems from the conventional paradigm of fixed-parameter models that cannot adapt to variations in test data (e.g., dose levels or scanner types) after training. To overcome this limitation and achieve robust generalization, we introduce U-TTT, a novel U-shaped model that integrates Test-Time Training (TTT) layers to dynamically adjust model parameters during inference through self-supervision, thereby adapting to the specific characteristics of each test instance. Furthermore, to comprehensively capture the complex degradations of 3D PET data, U-TTT features a dual-domain adaptation mechanism comprising a Spatial Test-Time Training (S-TTT) layer and a Frequency Test-Time Training (F-TTT) layer. The S-TTT layer captures and corrects spatial structural degradations, while the F-TTT layer suppresses global noise spectra and restores delicate high-frequency details. Extensive experiments demonstrate that U-TTT achieves state-of-the-art PET denoising performance and exhibits superior generalization under challenging distribution shifts, including both unseen dose levels and unseen scanners. Our code will be available at \href{https://github.com/Yaziwel/U-TTT.git}{https://github.com/Yaziwel/U-TTT}.


\end{abstract}
\section{Introduction} 

Positron Emission Tomography (PET) image denoising aims at recovers a high-quality full-dose PET image from its noisy low-dose counterpart. While deep learning-based methods  (e.g., CNNs \cite{xiang2017auto-contextcnn,chan2018dcnn,wang20183dcgan,luo2022argan,zhou2022sgsgan,yang2026unipet}
, Transformers \cite{jang2023spach,yang2023drmc,zeng20223d_cvtgan}, Mamba \cite{huang2025enhancing_mamba,chan2025dsamamba}, and RWKV \cite{yang2024restore_rwkv} models) have achieved remarkable success in PET image denoising, they predominantly rely on models with fixed parameters after training. This \textit{static} paradigm assumes that the testing data shares the same distribution as the training data. However, when deployed in unseen scenarios with variations in scanner and tracer dose levels, these fixed models often suffer from performance degradation due to distribution shift \cite{yu20253dddpm}. This limitation in generalization fundamentally restricts their robust deployment in real-world clinical applications.

Test-Time Training (TTT) \cite{sun2020ttt_rotation,gandelsman2022ttt_mae} has emerged as a promising paradigm for improving generalization under distribution shift. In conventional TTT frameworks, an auxiliary self-supervised task (e.g., rotation prediction \cite{sun2020ttt_rotation} or image reconstruction \cite{gandelsman2022ttt_mae}) is introduced alongside the primary task, and the model is trained jointly on both tasks. At test time, the model adapts to each test instance by updating its model parameters using only the auxiliary task objective. This per-sample optimization enables input-specific self-supervised adaptation and has been shown to improve robustness to distribution shift. However, a critical limitation arises from the misalignment between the auxiliary and primary objectives. Since the correlation between the two tasks is not explicitly enforced during the test-time update, minimizing the auxiliary loss does not guarantee an improvement in the primary task. In severe cases, this discrepancy can lead to overfitting the auxiliary task and catastrophic forgetting of the primary task.

To resolve this dilemma, recent works have introduced TTT layers \cite{sun2024ttt_rnn,zhou2024ttt_unet}, which reformulate the self-supervised adaptation process as an intrinsic component of the network architecture. Rather than treating the auxiliary task as an external procedure, TTT layers embed the auxiliary task’s parameter-update rule into the network’s forward pass so the auxiliary objective explicitly serves the primary task. Concretely, TTT layers follow a feature-level \textit{learn-and-adapt} paradigm. An inner model is trained on a feature self-reconstruction auxiliary task at test time, dynamically updating its parameters to adapt to the specific characteristics of the input prior to the final prediction. Crucially, because these inner-model updates are fully differentiable and executed within the global computational graph, the entire procedure can be meta-learned end-to-end using only the primary objective during training. This enforces an explicit alignment between the auxiliary and primary objectives, ensuring that: (1) the inner model learns from the auxiliary task and dynamically adjusts its parameters to adapt to the specific input; and (2) the outer model learns to align the auxiliary task with the primary objective, ensuring the adaptation fundamentally benefits the primary task. However, research on TTT layers is still in its early stages. Current literature \cite{sun2024ttt_rnn} predominantly investigates their efficiency in modeling long contexts—often positioning them as efficient alternatives to self-attention—while largely overlooking their potential to enhance generalization under distribution shifts, a critical requirement for robust PET image denoising. Moreover, because the TTT layer \cite{sun2024ttt_rnn} originated in natural language processing for 1D sequential data, a naive application to 3D vision tasks like PET denoising leads to suboptimal performance, necessitating task-specific design modifications.

In this paper, we introduce U-TTT, a U-shaped generalizable backbone designed for robust PET image denoising under diverse distribution shifts. To achieve optimal generalization, we argue that each test image defines a unique learning problem with its own generalization target. Our core innovation lies in the integration of Test-Time Training (TTT) layers, which dynamically update and adapt the model to each individual image during inference. However, existing TTT layers typically operate solely in the time (spatial) domain \cite{sun2024ttt_rnn,zhou2024ttt_unet,han2026vittt}, ignoring the complex, globally distributed noise present in the frequency domain. To overcome this, U-TTT performs dual-domain self-supervised adaptation: it employs a Spatial TTT (S-TTT) layer to capture and correct spatial structural degradations, alongside a Frequency TTT (F-TTT) layer to suppress global noise spectra while restoring high-frequency details. We further adapt the inner-model design and optimization to successfully scale both TTT layers from 1D sequences to 3D vision tasks. Consequently, the synergy between S-TTT and F-TTT enables the model to effectively address degradations and recover delicate details from both domains, achieving robust and high-quality PET image denoising. Extensive experiments show that U-TTT outperforms existing state-of-the-art methods for PET image denoising and generalizes best to unseen scanners and dose levels. 

Our contributions are threefold: 
\begin{itemize} 
\item We propose U-TTT, a novel U-shaped backbone that integrates Test-Time Training (TTT) layers to enable per-instance self-supervised adaptation during inference, effectively addressing the generalization limitations of static models under distribution shift in PET image denoising.
\item We innovatively introduce S-TTT and F-TTT layers for dual-domain self-supervised adaptation. This allows the model to comprehensively \textit{learn-and-adapt} to input characteristics from both spatial and frequency domains, leading to more comprehensive noise suppression and structural recovery.
\item Extensive experiments show that U-TTT outperforms state-of-the-art methods in PET image denoising and achieves superior generalization to unseen scanners and dose levels.
\end{itemize} 

\section{Method} 
\subsection{Overall Architecture} 
\begin{figure*}[t]
	\centering
	\includegraphics[width=\textwidth]{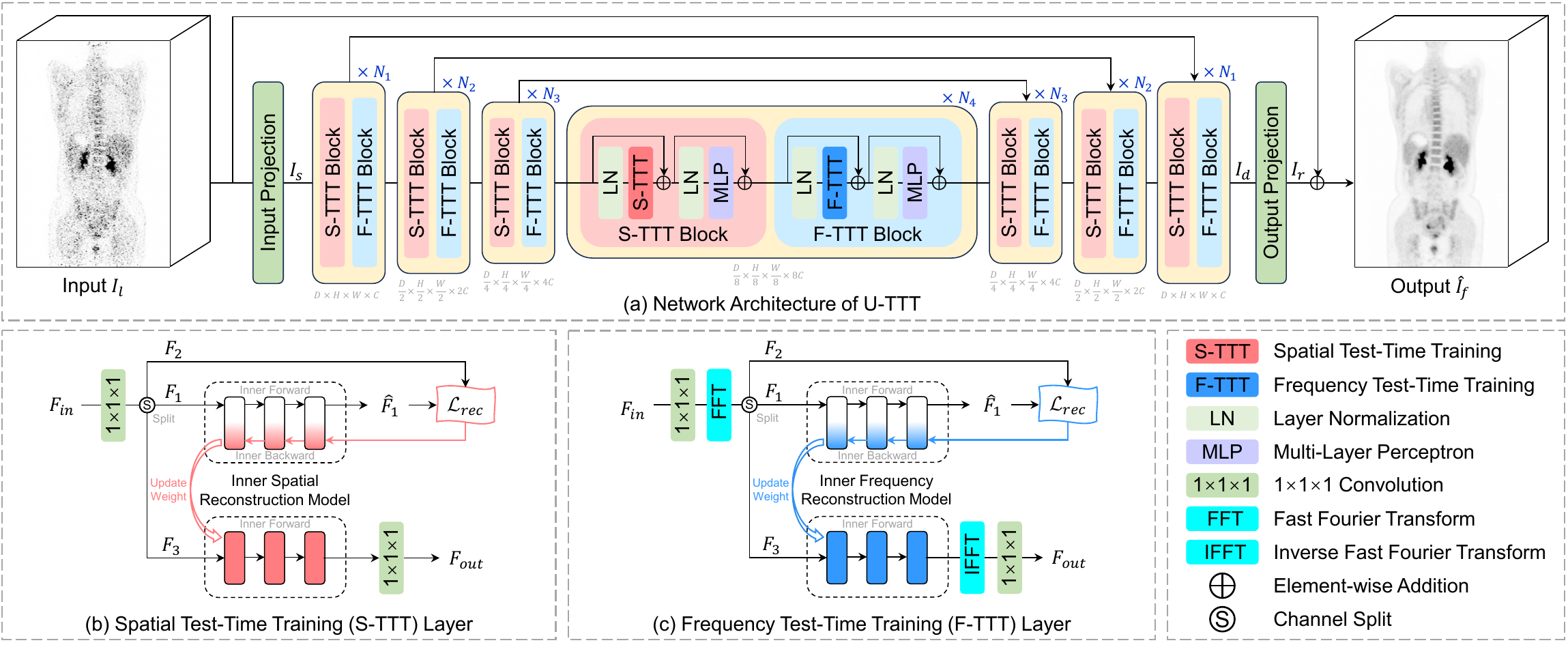}
    \caption{Overview of the proposed U-TTT.}
	\label{fig:framework}
\end{figure*}

Fig.~\ref{fig:framework} (a) presents the overall architecture of U-TTT, which aims to learn a robust model that recovers a high-quality full-dose PET image $\hat{I}_{f}$ from a given low-quality low-dose PET image $I_{l}$. Specifically, given a low-dose PET image $I_{l} \in \mathbb{R}^{D \times H \times W \times 1}$, U-TTT first extracts shallow features $I_{s} \in \mathbb{R}^{D \times H \times W \times C}$ using a $3 \times 3 \times 3$ convolutional input-projection layer, where $D \times H \times W$ represent the spatial dimensions and $C$ denotes number of channels. Next, these shallow features $I_{s}$ pass through a 4-level encoder-decoder U-shaped network and is transformed into a deep feature $I_{d} \in \mathbb{R}^{D \times H \times W \times C}$. To promote model generalization, each level of the encoder–decoder comprises consecutive Spatial Test-Time Training (S-TTT) and Frequency Test-Time Training (F-TTT) blocks for feature extraction. These blocks enable the model to dynamically update its parameters at test time and adapt to the test data, thereby improving generalizability. Finally, a $3 \times 3 \times 3$ output-projection convolutional layer transforms deep feature $I_{d}$ into a residual image $I_{r} \in \mathbb{R}^{D \times H \times W \times 1}$, which is added to the original low-dose image $I_{l}$ to yield the restored output $\hat{I}_{f} = I_{l} + I_{r}$. Both S-TTT and F-TTT blocks follow the macro architecture of a standard Transformer block, with the self-attention mechanism replaced by the respective S-TTT or F-TTT layer. We introduce S-TTT and F-TTT layers in detail in the subsequent sections.

\subsection{Spatial Test-Time Training Layer} 
The goal of the Spatial Test-Time Training (S-TTT) layer is to \textit{learn-and-adapt} in the spatial domain by modeling the spatial characteristics of the input. It performs a spatial feature reconstruction task to update an inner spatial reconstruction model, and then applies this adapted inner model to perform input-specific refinement of the features. The schematic of the S-TTT layer is shown in Fig.~\ref{fig:framework} (b). Given an input feature $F_{in} \in \mathbb{R}^{D \times H \times W \times C}$, S-TTT applies a $1 \times 1 \times 1$ convolution and splits the expanded channels to yield its three observations: 
\begin{equation}
    [F_1, F_2, F_3] = \text{ChannelSplit}(\text{Conv}_{1\times1\times1}(F_{in})),
\end{equation}
where $F_1, F_2, F_3 \in \mathbb{R}^{D \times H \times W \times C}$ serve as the input, target, and test feature for the spatial reconstruction task, respectively. To model the input-specific spatial characteristics, an inner spatial reconstruction model (SRM) with weights $W$ performs a $F_1 \to F_2$ reconstruction, with input $F_1$ and output $\hat{F}_1$: 
\begin{equation}
    \hat{F}_1 = \text{SRM}(W; F_1).
\end{equation}

The inner model design is crucial as its capacity determines how much it can learn from the input. Conventional TTT layers \cite{sun2024ttt_rnn,zhou2024ttt_unet} from NLP often use linear layers or MLPs that process tokens independently and thus fail to capture the structural connectivity required for 3D imaging. We therefore propose an efficient spatial reconstruction model tailored for 3D vision. It uses a spatial relation module (SRM) that processes different channel groups separately. Specifically, the input tensor $F_1$ is split into two subsets: the first $P$ channels ($F_1^{0:P}$) are processed by a $3\times3\times3$ depthwise convolution (DWConv), while the remaining $(C-P)$ channels ($F_1^{P:C}$) are transformed via a modified gated linear unit. The features from both branches are then concatenated along the channel dimension to yield the final representation:
\begin{equation}  
    \operatorname{SRM}(F_1) = \operatorname{Concat}(\left[ \operatorname{DWConv}(F_1^{0:P}) , \operatorname{FC}_1(F_1^{P:C}) \odot \operatorname{SiLU} \big(\operatorname{FC}_2(F_1^{P:C})\big) \right]),
\end{equation}

The inner SRM is then updated by minimizing a dot-product reconstruction loss $\mathcal{L}_{rec}(\hat{F}_1, F_2)=-\langle \hat{F}_1, F_2 \rangle$ between $\hat{F}_1$ and $F_2$ via online gradient descent:
\begin{equation}
W^* = W - \eta \frac{\partial \mathcal{L}_{rec}(\hat{F}_1, F_2)}{\partial W},
\end{equation}
where $\eta$ is the learnable inner model learning rate \cite{sun2024ttt_rnn} and $W^*$ denotes the updated weight. The optimized SRM is applied to perform tailored spatial processing on the test feature $F_3$ and the output feature $F_{out}$ is obtained by a $1 \times 1 \times 1$ convolution:
\begin{equation}
    F_{out} = \text{Conv}_{1\times1\times1}( \text{SRM}(W^*; F_3)).
\end{equation}

In this process, the spatial feature reconstruction $F_1 \to F_2$ is not an end objective, but rather a proxy mechanism for dynamically updating parameters and adapting the inner SRM to the characteristics of each input. Crucially, the entire inner-optimization loop—including the forward prediction, loss computation, and the gradient-based weight update—is fully differentiable and embedded as an unrolled computational graph within the outer denoising model. This design explicitly ensures that the outer model meta-learns to align this self-supervised auxiliary task with the primary denoising objective, thereby fundamentally preventing the objective misalignment that plagues conventional TTT paradigms.

\subsection{Frequency Test-Time Training Layer} 
Complementing the S-TTT layer, the Frequency Test-Time Training (F-TTT) layer, as illustrated in Fig.~\ref{fig:framework} (b), aims to \textit{learn-and-adapt} in the frequency domain. While S-TTT focuses on learning local spatial structures, F-TTT is specifically designed to model global frequency characteristics and suppress distributed noise spectra—a capability reported to be crucial for recovering delicate PET image details \cite{luo2022argan}. Although it shares an identical inner-optimization mechanism with S-TTT, its architecture is fundamentally tailored for spectral processing.

Given the input feature $F_{in}$, F-TTT first applies a $1\times1\times1$ convolution, followed immediately by a Fast Fourier Transform (FFT) to project the features into the frequency domain. The expanded spectral representations are then evenly split into three distinct observations, $F_1$, $F_2$, and $F_3$:
\begin{equation}
    [F_1, F_2, F_3] = \text{ChannelSplit}(\text{FFT}(\text{Conv}_{1\times1\times1}(F_{in}))).
\end{equation} 

The \textit{learn-and-adapt} process operates entirely within the frequency domain using an inner frequency reconstruction model (FRM). The architecture of FRM intentionally differs from its spatial counterpart. As each point in the frequency domain inherently encodes global spatial information, local operations such as depthwise convolutions become redundant. Consequently, FRM solely employs the modified gated linear unit for token-level spectral transformations:
\begin{equation}
    \hat{F}_1 = \text{FRM}(W; F_1) = \text{FC}(F_1) \odot \text{SiLU}(\text{FC}(F_1)).
\end{equation} 

Similar to S-TTT, the inner model FRM is dynamically optimized by minimizing the proxy reconstruction loss between the prediction $\hat{F}_1$ and the target $F_2$ via gradient descent to obtain the adapted weights $W^*$. Once adapted, the inner model performs tailored spectral modulation on the test feature $F_3$. Finally, an Inverse Fast Fourier Transform (IFFT) is applied to map the refined spectral features back to the spatial domain, followed by a concluding $1\times1\times1$ convolution to generate the final output:
\begin{equation}
    F_{out} = \text{Conv}_{1\times1\times1}( \text{IFFT}( \text{FRM}(W^*; F_3) ) ).
\end{equation}

With this design, F-TTT dynamically learns and adapts to the unique global spectral characteristics of each input. This enables the model to suppress distributed noise spectra while recovering fine high-frequency details, thereby complementing the structural refinements provided by the S-TTT layer. 

\begin{table}[t] 
\centering
\caption{Dataset information.} 
\resizebox{\textwidth}{!}{
\begin{tabular}{cc|c|c|c|c|c|c|c|c|c|c|c}
\toprule
\multicolumn{2}{c|}{\textbf{Dataset}}                     & Institution & Type       & Scanner & Tracer       & Administered Dose & DRF         & Spacing (mm$^3$)           & Shape                            & Train & Validation & Test \\ \midrule
\multicolumn{1}{c|}{Base   Dataset}               & $D_1$ & $I_1$       & Whole Body & $S_1$   & $^{18}$F-FDG & 371 MBq           & 2, 3, 6, 12 & 3.15 $\times$ 3.15 $\times$ 1.87 & 192 $\times$ 192 $\times slices$ & 90    & 10         & 30   \\ \hline
\multicolumn{1}{c|}{OOD-DRF}                      & $D_2$ & $I_1$       & Whole Body & $S_1$   & $^{18}$F-FDG & 353 MBq           & 4, 10       & 3.15 $\times$ 3.15 $\times$ 1.87 & 192 $\times$ 192 $\times slices$ & -     & -          & 30   \\ \hline
\multicolumn{1}{c|}{\multirow{2}{*}{OOD-Scanner}} & $D_3$ & $I_2$       & Whole Body & $S_2$   & $^{18}$F-FDG & 347 MBq           & 4           & 3.12 $\times$ 3.12 $\times$ 1.75 & 192 $\times$ 192 $\times slices$ & -     & -          & 30   \\ \cline{2-13} 
\multicolumn{1}{c|}{}                             & $D_4$ & $I_3$       & Whole Body & $S_3$   & $^{18}$F-FDG & 221 MBq           & 4           & 1.65 $\times$ 1.65 $\times$ 1.65 & 256 $\times$ 256 $\times slices$    & -     & -          & 30   \\ \bottomrule
\end{tabular}
}
\label{tab:dataset}
\end{table}  

\subsection{Loss Function} 
Following prior work on PET image denoising \cite{zeng20223d_cvtgan}, the total loss function $\mathcal{L}_{total}$ is formulated as a weighted sum of an $\mathcal{L}_1$ loss—which enforces accurate reconstruction of image content—and a generative adversarial loss $\mathcal{L}_{adv}$—which promotes fine detail recovery through adversarial learning \cite{goodfellow2020generative}:
\begin{equation}
    \mathcal{L}_{total} = \mathcal{L}_1 + \lambda\mathcal{L}_{adv},
\end{equation} 
where the balancing factor is set to $\lambda = 1 \times 10^{-3}$.

\section{Experiments and Results} 
\subsection{Dataset} 
To demonstrate the effectiveness and generalizability of the proposed U-TTT, we establish four distinct whole-body PET datasets ($D_1$–$D_4$) with diverse characteristics, as summarized in Tab.~\ref{tab:dataset}. For each dataset, we first collect full-dose PET data from patients in list-mode. Corresponding low-dose PET data are then simulated by randomly downsampling the list-mode data according to a predefined dose reduction factor (DRF) (e.g., retaining 25\% of the data for a DRF of 4). Both full- and low-dose PET images are subsequently reconstructed from the list-mode data using the standard OSEM algorithm \cite{hudson1994osem}. Note that institutional and scanner identities have been anonymized. The base dataset $D_1$ is utilized for model training, validation, and in-distribution testing. The remaining datasets are reserved for out-of-distribution (OOD) testing: $D_2$ evaluates performance on previously unseen DRFs, while $D_3$ and $D_4$ assess generalizability across previously unseen scanners. During training, images are split into 3D patches of size $64\times64\times64$. At test time, the full estimated PET image is reconstructed by stitching the predicted patches together. 

\begin{table}[t] 
\centering
\caption{Comparison results on the in-distribution base dataset.} 
\resizebox{\textwidth}{!}{
\begin{tabular}{c|c|c|ccccc|ccccc|ccccc}
\toprule
\multirow{2}{*}{\textbf{Method}} & \multirow{2}{*}{Params (M)} & \multirow{2}{*}{FLOPs (G)} & \multicolumn{5}{c|}{PSNR↑}                                                         & \multicolumn{5}{c|}{SSIM↑}                                                              & \multicolumn{5}{c}{Lesion Error↓}                                                       \\ \cline{4-18} 
                                 &                             &                            & DRF=2          & DRF=3          & DRF=6          & DRF=12         & Avg.           & DRF=2           & DRF=3           & DRF=6           & DRF=12          & Avg.            & DRF=2           & DRF=3           & DRF=6           & DRF=12          & Avg.            \\ \midrule
3D-cGAN \cite{wang20183dcgan}                          & 53.62                       & 12.11                      & 49.87          & 48.61          & 46.88          & 44.51          & 47.47          & 0.9772          & 0.9696          & 0.9564          & 0.9366          & 0.9600          & 0.1451          & 0.1624          & 0.1987          & 0.2628          & 0.1923          \\
DRMC  \cite{yang2023drmc}                           & 0.54                        & 137.38                     & 50.11          & 48.96          & 47.13          & 44.63          & 47.71          & 0.9783          & 0.9701          & 0.9553          & 0.9391          & 0.9607          & 0.1496          & 0.1687          & 0.2057          & 0.2769          & 0.2002          \\
Spach Transformer  \cite{jang2023spach}              & 19.22                       & 163.38                     & 50.52          & 49.18          & 47.21          & 44.79          & 47.93          & 0.9766          & 0.9709          & 0.9571          & 0.9408          & 0.9614          & 0.1352          & 0.1528          & 0.1943          & 0.2555          & 0.1845          \\
3D DDPM \cite{yu20253dddpm}                         & 209.63                      & 2600444.69                 & 50.23          & 48.71          & {\ul 47.32}    & 44.87          & 47.78          & 0.9791          & 0.9712          & 0.9588          & 0.9357          & 0.9612          & 0.1152          & 0.1372          & 0.1834          & 0.2437          & 0.1699          \\
VQPET   \cite{chen2026vqpet}                         & 106.34                      & 439.99                     & {\ul 50.72}    & {\ul 49.31}    & 47.31          & {\ul 45.08}    & {\ul 48.11}    & {\ul 0.9801}    & {\ul 0.9723}    & {\ul 0.9595}    & {\ul 0.9451}    & {\ul 0.9643}    & {\ul 0.1107}    & {\ul 0.1291}    & {\ul 0.1801}    & {\ul 0.2359}    & {\ul 0.1640}    \\
U-TTT (Ours)                            & 10.20                       & 43.52                      & \textbf{51.63} & \textbf{50.01} & \textbf{48.02} & \textbf{45.96} & \textbf{48.91} & \textbf{0.9823} & \textbf{0.9741} & \textbf{0.9621} & \textbf{0.9498} & \textbf{0.9671} & \textbf{0.0903} & \textbf{0.1178} & \textbf{0.1583} & \textbf{0.2278} & \textbf{0.1486} \\ \bottomrule
\end{tabular}
}
\label{tab:id_test}
\end{table}  

\begin{figure*}[t]
	\centering
	\includegraphics[width=\textwidth]{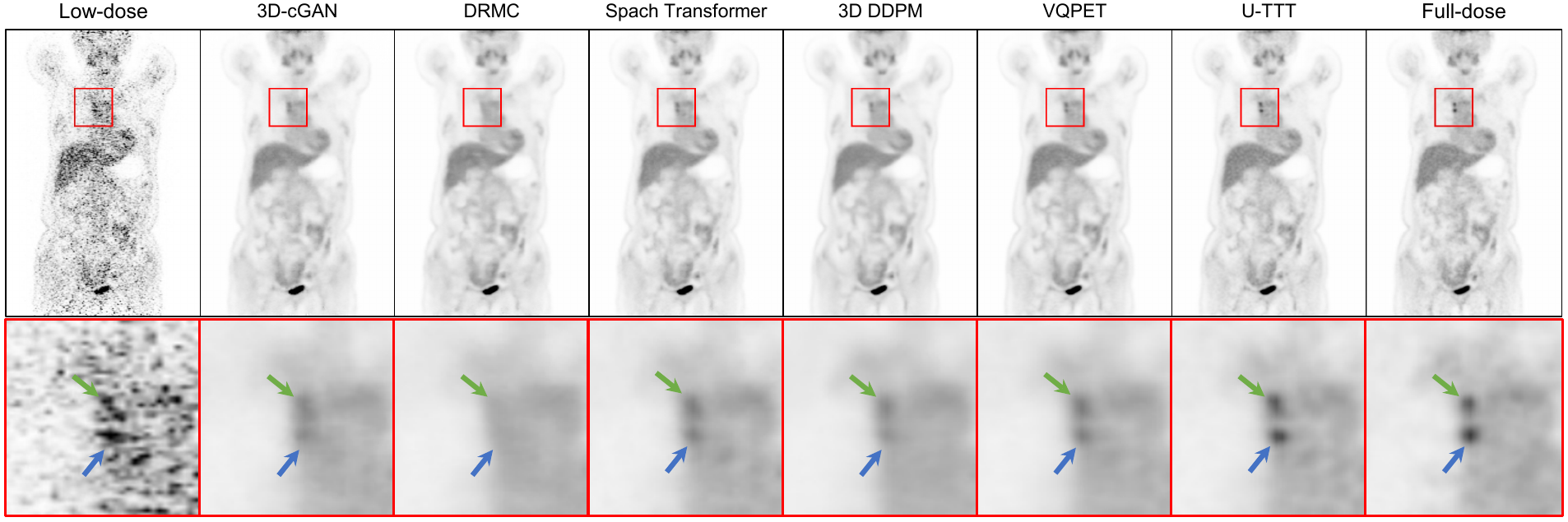}
    \caption{Visualization comparison on the in-distribution base dataset at DRF=12.}
	\label{fig:comparison}
\end{figure*} 

\subsection{Implementation} 
For the model architecture, the number of feature extraction blocks in U-TTT are set to $N_1=2$, $N_2=4$, $N_3=6$, and $N_4=8$, and the input projection channel dimension is $C=24$. In the inner model, the number of channels processed by the DWConv is set to $P=24$. For training, we use a batch size of 4 and optimize the model with the AdamW optimizer at an initial learning rate of $1\times10^{-4}$ for $3\times10^{5}$ iterations. For model evaluation, we choose PSNR and SSIM to evaluate the overall image quality. For clinical evaluation, we quantify the standardized uptake value (SUV) errors within lesion regions using mean absolute error.

\subsection{Comparative Experiments}
We compare our U-TTT with other five state-of-the-art PET image denoising methods: the GAN-based 3D-cGAN \cite{wang20183dcgan}, the Transformer-based DRMC \cite{yang2023drmc} and Spach Transformer \cite{jang2023spach}, the diffusion-based 3D DDPM \cite{yu20253dddpm}, and the vector quantization (VQ) codebook prior-based VQPET \cite{chen2026vqpet}. All methods are trained on the base dataset $D_1$ and tested on all four dataset ($D_1$-$D_4$) to evaluate both in-distribution and out-of-distribution performance.

\textbf{Comparison Results on the In-Distribution Dataset.} Tab.~\ref{tab:id_test} reports results on the base dataset $D_1$. U-TTT outperforms all comparison methods at four different DRFs across three metrics while maintaining good computational efficiency. Notably, U-TTT surpasses the second-best VQPET by an average of 0.80 dB in PSNR and 0.0028 in SSIM, while reducing lesion error by 0.0154. These results indicate that U-TTT achieves superior denoising performance, delivering high overall quantitative accuracy and improved recovery of small lesions. As shown in Fig.\ref{fig:comparison}, U-TTT effectively restores the contrast of two small lesions, whereas other methods suffer from over-smoothing.

\textbf{Comparison Results on Out-Of-Distribution Datasets.} We evaluate model generalizability against distribution shifts using three out-of-distribution (OOD) datasets: the OOD-DRF dataset ($D_2$) featuring previously unseen DRFs, and the OOD-Scanner datasets ($D_3$ and $D_4$) acquired from unseen scanners. Results in Tab.~\ref{tab:ood_test} indicates that U-TTT achieves the best performance when handling distribution shift on dose levels and scanners. This superior generalization capability is directly attributed to U-TTT's \textit{learn-and-adapt} mechanism, which empowers the model to dynamically adjust its parameters to the specific characteristics of each test instance during inference.

\begin{table}[t] 
\centering
\caption{Comparison results on the out-of-distribution datasets.} 
\resizebox{\textwidth}{!}{
\begin{tabular}{c|c|c|cccccc|cccccc}
\toprule
\multirow{3}{*}{\textbf{Method}} & \multirow{3}{*}{Params (M)} & \multirow{3}{*}{FLOPs (G)} & \multicolumn{6}{c|}{OOD-DRF}                                                                                                & \multicolumn{6}{c}{OOD-Scanner}                                                                                             \\ \cline{4-15} 
                                 &                             &                            & \multicolumn{3}{c|}{PSNR↑}                                            & \multicolumn{3}{c|}{SSIM↑}                          & \multicolumn{3}{c|}{PSNR↑}                                            & \multicolumn{3}{c}{SSIM↑}                           \\ \cline{4-15} 
                                 &                             &                            & DRF=4          & DRF=10         & \multicolumn{1}{c|}{Avg.}           & DRF=4           & DRF=10          & Avg.            & $S_2$          & $S_3$          & \multicolumn{1}{c|}{Avg.}           & $S_2$           & $S_3$           & Avg.            \\ \midrule
3D-cGAN \cite{wang20183dcgan}                         & 53.62                       & 12.11                      & 46.94          & 44.18          & \multicolumn{1}{c|}{45.56}          & 0.9602          & 0.9371          & 0.9487          & 39.78          & 43.32          & \multicolumn{1}{c|}{41.55}          & 0.9318          & 0.9462          & 0.9390          \\
DRMC  \cite{yang2023drmc}                           & 0.54                        & 137.38                     & 46.71          & 44.36          & \multicolumn{1}{c|}{45.54}          & 0.9621          & 0.9332          & 0.9477          & 39.92          & 43.51          & \multicolumn{1}{c|}{41.72}          & 0.8432          & 0.9021          & 0.8727          \\
Spach Transformer   \cite{jang2023spach}             & 19.22                       & 163.38                     & 47.23          & 44.49          & \multicolumn{1}{c|}{45.86}          & 0.9594          & 0.9356          & 0.9475          & {\ul 40.41}    & 43.91          & \multicolumn{1}{c|}{{\ul 42.16}}    & 0.9148          & 0.9265          & 0.9207          \\
3D DDPM  \cite{yu20253dddpm}                        & 209.63                      & 2600444.69                 & 47.45          & 44.58          & \multicolumn{1}{c|}{46.02}          & 0.9635          & 0.9384          & 0.9510          & 39.96          & 42.88          & \multicolumn{1}{c|}{41.42}          & 0.9326          & 0.9481          & 0.9404          \\
VQPET  \cite{chen2026vqpet}                          & 106.34                      & 439.99                     & {\ul 47.72}    & {\ul 44.66}    & \multicolumn{1}{c|}{{\ul 46.19}}    & {\ul 0.9648}    & {\ul 0.9418}    & {\ul 0.9533}    & 40.18          & {\ul 44.03}    & \multicolumn{1}{c|}{42.11}          & {\ul 0.9344}    & {\ul 0.9506}    & {\ul 0.9425}    \\
U-TTT (Ours)                     & 10.20                       & 43.52                      & \textbf{48.51} & \textbf{45.21} & \multicolumn{1}{c|}{\textbf{46.86}} & \textbf{0.9683} & \textbf{0.9481} & \textbf{0.9582} & \textbf{41.31} & \textbf{44.89} & \multicolumn{1}{c|}{\textbf{43.10}} & \textbf{0.9449} & \textbf{0.9578} & \textbf{0.9514} \\ \bottomrule
\end{tabular}
}
\label{tab:ood_test}
\end{table}

\begin{table*}[t]
\begin{floatrow}
\capbtabbox{ 
\resizebox{0.43\textwidth}{!}{
\begin{tabular}{c|ccc}
\toprule
\multirow{2}{*}{\textbf{Method}} & \multicolumn{3}{c}{PSNR↑}                                                                  \\ \cline{2-4} 
                                 & \multicolumn{1}{c|}{Base Dataset}   & \multicolumn{1}{c|}{OOD-DRF}        & OOD-Scanner    \\ \midrule
Baseline                         & \multicolumn{1}{c|}{47.63}          & \multicolumn{1}{c|}{45.68}          & 41.89          \\
S-TTT                            & \multicolumn{1}{c|}{48.45}          & \multicolumn{1}{c|}{46.07}          & 42.43          \\
F-TTT                            & \multicolumn{1}{c|}{{\ul 48.72}}    & \multicolumn{1}{c|}{{\ul 46.55}}    & {\ul 42.78}    \\
S-TTT + F-TTT (Ours)             & \multicolumn{1}{c|}{\textbf{48.91}} & \multicolumn{1}{c|}{\textbf{46.86}} & \textbf{43.10} \\ \bottomrule
\end{tabular}
}
}{
 \caption{Component analysis.}
 \label{tab:component_analysis}
}

\capbtabbox{ 

\resizebox{0.40\textwidth}{!}{ 
\begin{tabular}{c|ccc}
\toprule
\multirow{2}{*}{\textbf{Inner   Model}}      & \multicolumn{3}{c}{PSNR↑}                                                                  \\ \cline{2-4} 
                                             & \multicolumn{1}{c|}{Base Dataset}   & \multicolumn{1}{c|}{OOD-DRF}        & OOD-Scanner    \\ \midrule
Linear   \cite{sun2024ttt_rnn}                                    & \multicolumn{1}{c|}{48.31}          & \multicolumn{1}{c|}{45.97}          & 42.21          \\
MLP   \cite{sun2024ttt_rnn}                                       & \multicolumn{1}{c|}{48.57}          & \multicolumn{1}{c|}{46.15}          & 42.54          \\
FC(x)$\odot$SiLU(FC(x))                      & \multicolumn{1}{c|}{{\ul 48.80}}    & \multicolumn{1}{c|}{{\ul 46.77}}    & {\ul 42.96}    \\
Ours & \multicolumn{1}{c|}{\textbf{48.91}} & \multicolumn{1}{c|}{\textbf{46.86}} & \textbf{43.10} \\ \bottomrule
\end{tabular}
}
}{
 \caption{Inner model design.}
 \label{tab:inner_design}
 \small
} 

\end{floatrow} 
\end{table*}

\subsection{Ablation Study} 
We perform ablation studies to validate the effectiveness of the core components in the S-TTT and F-TTT layers. As shown in Tab.~\ref{tab:component_analysis}, we establish a baseline model by replacing these layers with MLP layers. The introduction of either the S-TTT or F-TTT layer yields significant improvements over the baseline, with the F-TTT layer proving more effective than S-TTT. The combination of both layers achieves the best overall performance. We also investigate the effectiveness of our inner model design. As reported in Tab.~\ref{tab:inner_design}, utilizing the modified efficient gated linear unit within the S-TTT and F-TTT layers significantly outperforms conventional linear and MLP layers \cite{sun2024ttt_rnn}. Furthermore, incorporating depth-wise convolution into the S-TTT layer results in additional performance gains.

\section{Conclusion}
We propose U-TTT, a novel PET denoising framework that leverages Test-Time Training to enable dynamic model adaptation during inference. By incorporating dual-domain S-TTT and F-TTT layers, the model effectively learns instance-specific characteristics from both spatial and frequency domains to restore structural details and suppress global noise.  Extensive experiments demonstrate that U-TTT achieves state-of-the-art performance and superior generalizability.
\bibliographystyle{splncs04}
\bibliography{ref}

\end{document}